\documentclass{article}

\usepackage{arxiv}
\usepackage{caption}
\usepackage{graphicx}
\usepackage[utf8]{inputenc} 
\usepackage[T1]{fontenc}    
\usepackage{hyperref}       
\usepackage{url}            
\usepackage{booktabs}       
\usepackage{amsfonts}       
\usepackage{nicefrac}       
\usepackage{microtype}      
\usepackage{lipsum}

\title{Aspect-Based Sentiment Analysis in Education Domain}

\author{
  Rinor Hajrizi, Krenare Pireva Nuçi  \\ 
  Department of Computer Science and Engineering\\
  University for Business and Technology\\
  10000 Prishine, Kosovo \\
  \texttt{rh40288@ubt-uni.net, krenare.pireva@ubt-uni.net} \\

}

\begin{document}
\maketitle

\begin{abstract}
Analysis of a large amount of data has always brought value to institutions and organizations. Lately, people's opinions expressed through text have become a very important aspect of this analysis. In response to this challenge, a natural language processing technique known as Aspect-Based Sentiment Analysis (ABSA) has emerged. Having the ability to extract the polarity for each aspect of opinions separately, ABSA has found itself useful in a wide range of domains. Education is one of the domains in which ABSA can be successfully utilized. Being able to understand and find out what students like and don't like most about a course, professor, or teaching methodology can be of great importance for the respective institutions. While this task represents a unique NLP challenge, many studies have proposed different approaches to tackle the problem. In this work, we present a comprehensive review of the existing work in ABSA with a focus in the education domain. A wide range of methodologies are discussed and conclusions are drawn.
\end{abstract}


\section{Introduction}
To understand the meaning and goal of Aspect-Based Sentiment Analysis, we should first familiarize ourselves with Sentiment Analysis. Sentiment Analysis refers to the use of Natural Language Processing (NLP) and text analysis techniques to perform the task of identifying, extracting, and studying subjective information \cite{liu2012sentiment}. It is also known as Opinion Mining because what it does is extract the opinion given by a relatively large audience in a particular domain. This opinion could be referring to customer reviews on a product, comments about a particular subject in social media, feedback given by participants of a particular experience, and what interests us it is also the opinion of the students toward teacher, course, and Institution. People usually care about the opinions expressed by others, because they represent value for them as individuals or mostly for their institutions and organizations. Their value becomes larger, the larger the number of opinions expressed. A problem that arises in this situation corresponds to the limited ability of the owners of data to translate them into value. In other words, it is impossible (or extremely inefficient) for humans to sit down and analyze a whole corpus of data and eventually come up with a conclusion on whether the sentiment expressed was positive, negative, or neutral (a process which is also known as sentiment polarity analysis). Fortunately, in parallel with the advancement of technology, several researchers are proposing new techniques that could automatically extract sentiment polarities from a given opinion. The solutions proposed usually employ machine learning techniques and algorithms, which have gained huge popularity lately. In addition to the initial challenge, we are sometimes interested in more detailed insights concerning the expressed sentiment.  
Today, having performed only sentiment polarity analysis into a particular domain could not be encountered as enough, when it comes to fulfilling people's and their organizations' needs. In this regard, many times, we need to have more insight, for example: having the evaluation of students toward their teacher or course, in many cases the teachers want to know what aspects are discussed once the student gave their opinion. Sometimes teachers or higher management needs to know what aspect of the subject was evaluated positively and what aspect negatively. This advancement of sentiment analysis, where people need to deal with sentiments related to each aspect of importance, today is encountered as a new branch of Sentiment Analysis, namely Aspect-Based Sentiment Analysis.  
This paper aims to offer a comprehensive review and analysis of state-of-the-art research papers that are conducted so far in the field of Aspect-Based Sentiment Analysis. The outcome of this report is structured in the sections that follow, with the utmost goal of providing a better perspective to the field, and possibly end up with conclusions based on collected facts. The rest of the paper is organized as follows: Section 2 introduces us to the steps, procedures, and practices that have been witnessed in the existing work, followed by Section 3 which focuses on the datasets used and the domains that have found proper applications of the techniques that are discussed and compared in detail in section 5. Before reaching the Sentiment Analysis techniques, we should get familiar with the techniques employed in data preprocessing and the importance of this phase in the overall success, which can be found in section 4. Finally, the paper concludes in section 6, where a final opinion is drawn and lessons learned are explicitly stated.

\section{Steps, Procedures and Practices in Aspect-Based Sentiment Analysis }
\label{sec:steps-procedures}

\begin{figure}[h]
    \begin{center}
        \includegraphics[width=120mm]{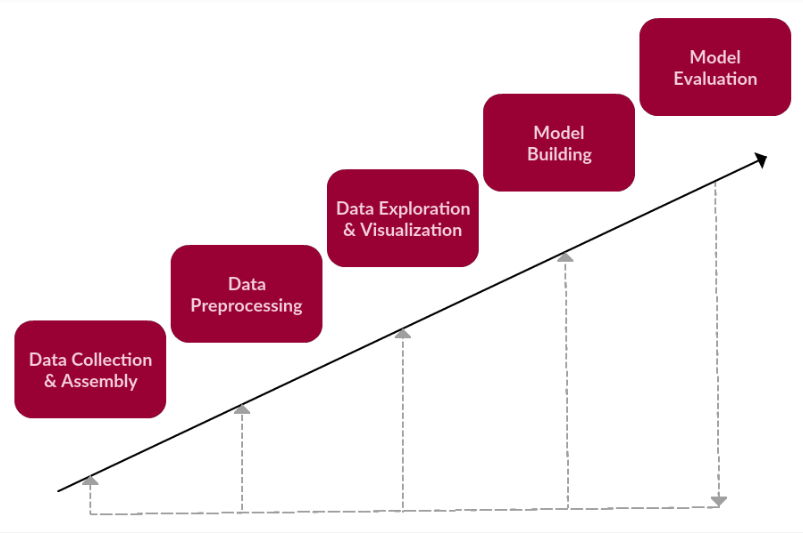}
        \caption{Text mining processes \cite{mayo2017general}}
        \label{fig:fig1}
    \end{center}
\end{figure}

During the review of the existing work, there is identified a common pattern of steps and procedures that are followed for performing aspect-based sentiment analysis. Initially, while performing text mining, the common processes that need to be followed are depicted in Figure \ref{fig:fig1}. These include Data collection or Assembly, Data Preprocessing, Data Exploration and Visualization, Model Building, and Model Evaluation.

Initially, each process within Figure \ref{fig:fig1} is mentioned as a general concept, and later all implementation procedures are explained in detail.  Even though the process seems to be linear in theory, in practice it is rather iterative, with many kinds of variations. The data collection process, as the first step, refers to the process of gathering all necessary data that are needed for extracting the intended knowledge that helps us to draw the main conclusions at the end. In text mining, this usually means gathering the necessary text data for the particular domain that is generated by human entities. In this respect, starting with data gathering from various sources, with context to student evaluation processes, feedback sessions, web scraping, review sections, and many other sources that can help in the domain of the defined problem that we are eager to analyze.  
The second step (see Figure \ref{fig:fig1}), the data preprocessing phase, refers to the performed activities that will bring the text data in the desired shape. Usually, the text data that is collected from various sources carries a considerable amount of noise and non-sense symbols that is necessary and meaningful while residing in the initial habitat, but eventually not necessary or even degrading while trying to extract value out of it.  
Further, data exploration and visualization as a next step try to create a sense of understanding of the data that we are dealing with. In other words, it attempts to raise a level of intuition and thus sharpen our thought and probably affect our decisions made in the next steps. We can for example notice a strange pattern in the text data that we were given, a higher abundance of a particular type of opinion, and a lower one for another type of opinion. We might sense that a dataset is imbalanced by simply exploring the data manually or visualizing it with automated tools.  
Next, we reach one of the most important steps, namely the model building process. This is where the author intends to design and build the model including a number of algorithms and frameworks using clean or processed data. If a machine learning approach is followed, this step constitutes of choosing the appropriate ML model, deciding about its parameters, and specific algorithms to be used. Having our model built, we are left with the last and most rewarding step: model evaluation. This is where an experiment is conducted. The collected and preprocessed data is fed to the built model, and results are generated. Those results are analyzed and compared with the results from other approaches or models using various techniques, and a conclusion on whether the chosen approach was successful or not, has met our hypothesis, or whether the final result is enough to answer the raised research questions. Depending on the case, the first and second steps (rarely also the third one) in many cases are generated automatically using existing tools and its libraries, or even sometimes are skipped at all. The latter situation happens usually when the researchers are working on a public dataset that already covers the structured data that needs to be used. In the next session, we are going to discuss in more detail regarding these datasets, and its application domains where Sentiment Analysis (hereafter: SA) specifically Aspect-Based Sentiment Analysis (hereafter: ABSA) techniques have been effective.

\section{Datasets and Application Domains }

While reviewing several existing research work, a wide range of domains are mentioned and treated. Today, for each of those domains different datasets of different sizes have made use of it. In this section, a classification of domains and the corresponding datasets are provided, and in some cases, a rationale behind it is also given. 
Today, there are a number of researches that include the data analysis in the: 

\begin{itemize}
\item Restaurants domain: \cite{xue2018aspect,ma2018targeted,pontiki2015semeval,ruder2016hierarchical, sun2019utilizing,wang2016recursive,zhang2019aspect,hazarika2018modeling,afzaal2019tourism, majumder2018iarm,mao2019aspect,mubarok2017aspect,nguyen2015phrasernn,akhtar2019language,tang2016aspect};
\item Technology: \cite{xue2018aspect, ma2018targeted, pontiki2015semeval, ruder2016hierarchical, sun2019utilizing, wang2016recursive, zhang2019aspect, hazarika2018modeling, majumder2018iarm, mao2019aspect, mubarok2017aspect, nguyen2015phrasernn, akhtar2019language, tang2016aspect, xu2019bert}; 
\item self-driving cars \cite{ahmad2017sentiment};  
\item hotels \cite{ruder2016hierarchical, afzaal2019tourism};
\item movies \cite{dos2014deep};  
\item phones and cameras \cite{ruder2016hierarchical};  
\item online-shopping \cite{ilmania2018aspect}, and 
\item online-courses \cite{kastrati2020weakly, itani2018understanding, whitehill2015beyond, kastrati2020wet, kastrati2020aspect};
\end{itemize}

The datasets that correspond to these domains came from different sources. While most of work conducted so far in restaurants and technology reviews came from the readily SemEval series of evaluations of computational semantic analysis systems, with readily labelled datasets, authors of other papers also used datasets from Sentihood \cite{ma2018targeted, sun2019utilizing}; Twitter \cite{dos2014deep, imran2020cross,9207881, zhang2019aspect, mao2019aspect, ahmad2017sentiment}; Amazon \cite{xu2019bert}; Yelp \cite{xu2019bert}; Coursera \cite{kastrati2020weakly}, online market \cite{ilmania2018aspect}, to name a few. Manual data collection was performed in \cite{afzaal2019tourism} using Web Crawler and APIs to collect restaurant and hotel reviews (2000 and 4000 respectively). As can be seen for the reviewed paper \cite{itani2018understanding}, which anticipates the dropout rates in Massive Open Online Courses (MOOC) and the reasons behind \cite{dalipi2018mooc, dalipi2016towards, imran2016analysis, pireva2015user}, the dataset consists of activity traces of 20,142 premium learners in OpenClassroom platform. More specifically, the activity traces were collected from “Create your website with HTML” and “Understanding the Web” courses. The dataset contains subscription related events, course related events, and exercise session events. From these activity traces, authors extracted features and then defined rules for particular activity trace combinations. Furthermore, in \cite{imran2019predicting} and evaluation of deep learning models was performed, while in \cite{whitehill2015beyond} an automatic intervention mechanism for MOOC student stop out was presented. In contrast to drop out, the stop out is defined as an action of temporal withdrawal from a course. And, in \cite{whitehill2015beyond} is addressed a stop out classifier, which is used to orchestrate an intervention before students dropped out, and to survey them dynamically about why they ceased participation. The analysis was conducted based on the dataset formed by collecting data from three batches of survey emails, sent to 5073 students. In addition to the existing work in the education domain and the corresponding datasets used, a dataset for this research paper was created manually as well. Feedback was requested from students who attended the online-only Human-Computer Interaction University course. Their feedback covered different aspects, such as: evaluation for the professor, for course, for the institution, for the online learning experience, for assessment methods, and project-related activities. Since the feedback of students was given in a document format, the data had to undergo a cleaning phase before proceeding in the analysis. In addition to pre-processing, the raw text is structured into XML format and manually annotated. The annotation represented the polarity expressed by students toward each category in sentence level. 
Since most of the reviewed work is oriented around supervised models, researchers have preferred to rely on readily annotated datasets, specifying whether a sentence or review is positive or negative and sometimes indicating the word or words that contributed to the overall polarity. In addition to the manual annotation of large datasets being a time-consuming activity, it is also prone to human error and thus wrong classification \cite{kastrati2020weakly}. In domains where readily annotated datasets were missing, authors chose to use unsupervised or weakly-supervised techniques for the sole reason of avoiding this bottleneck in their work \cite{kastrati2020weakly, dos2014deep}. In rare cases, authors have chosen to manually annotate the datasets and make use of supervised techniques. Having currently some ground facts and statements defined, we are ready to continue with the next two sections, in which we dwell on the core activities in ABSA, namely Preprocessing and Modelling for Sentiment Analysis.

\section{Data Preprocessing Techniques and Importance}
Data preprocessing as stated earlier brings the text in a desired shape and structure. Preprocessing of text was present in almost all papers that have been reviewed. They have approached the problem while applying different modifications in the data. The task of preprocessing can be thought of as a collection of three different activities:

\begin{itemize}
\item Tokenization,  
\item Normalization, and 
\item Noise removal \cite{mayo2017general}.  .
\end{itemize}

\begin{figure}[h]
    \begin{center}
        \includegraphics[width=90mm]{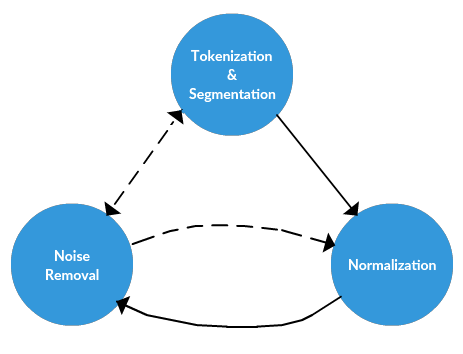}
        \caption{Text mining processes \cite{mayo2017general}}
        \label{fig:fig2}
    \end{center}
\end{figure}

A visual representation of the process is depicted in Figure \ref{fig:fig2}. As you can infer from the picture this process should not be encountered as linear, but iterative and with no clear ordering of activities. Tokenization as the first activity discussed here refers to the process of splitting longer strings of text into shorter ones, usually by following syntactic rules. In most cases, a whole document can be split into paragraphs, a paragraph into sentences, and a sentence into words. This process may seem straightforward, but sometimes it is not particularly clear when a sentence is finished by simply relying on punctuation. Further, the situation gets more complex when dealing with foreign languages that we cannot make sense of it. In these cases, it is not possible to proceed with tokenization without consulting someone who knows the particular language. This step is usually conducted so that it is easier and more manageable to play with the textual data, i.e. we can modify individual words, based on the existing knowledge that we have from the use of a dictionary. Otherwise, we modify the key points within the sentences, paragraphs, and documents. Sometimes tokenization goes up to the character level, even though very rarely \cite{mayo2017general}.  
Additionally, when analyzing Figure \ref{fig:fig2}, there is another important step in text preprocessing, and that is normalization. It represents a series of related tasks to put all text in the same level playing field. This might include \cite{mayo2017general}:

\begin{itemize}
\item   converting all text to the same case (uppercase or lowercase, mostly lowercase), 
\item  removing punctuation,  
\item  converting numbers to their word equivalents, to name a few. 

\end{itemize}

This allows processing to proceed uniformly throughout the data. Some more known steps that are undertaken during normalization are \cite{mayo2017general}: 

\begin{itemize}
    \item stemming,  
    \item lemmatization, and  
    \item removal of stop words.  
\end{itemize}

Stemming is the process of removing suffixes, prefixes, infixes, and circumfixes to obtain a word stem. For example, from a single word "playing" we obtain the root word "play", or from “computation” we obtain “compute”, and so on. On the other hand, lemmatization, even though related to stemming, can catch much more in a word. It attempts to bring the word in its canonical base form \cite{mayo2017general}. For example, from the word "felt" we can obtain "feel". Finally, since a lot of sentiment analysis techniques rely in a way or another on word count, stop-words removal is performed. Stop-word is a term for a word that is used very frequently in a language but provides no sentiment or value, thus nothing can be deduced from its high presence \cite{saif2014stopwords}. Stop-words are normally removed from the text and then further processing can proceed. Finally, as the last step in text data preprocessing (see Figure \ref{fig:fig2}), noise removal is conducted. It is a process whose actual activities are very much dependent on the task at hand. For example, after text data has been extracted from web pages, removing the HTML tags corresponds to noise removal. Greetings and introductory phrases, despite being a respectful gesture in the context of human interaction, are nothing more than noise and confusion to a text-processing algorithm. These and much more constitute to noise removal as the last step in data preprocessing that was witnessed during the literature review. In the next section, a thorough discussion of sentiment analysis techniques used is presented.

\section{ Aspect-Based Sentiment Analysis Techniques and Approaches (Model Building) }
\label{sec:aspect-based}

Most of the work reviewed has treated an Aspect-Based Sentiment Analysis challenge to be composed of two inner challenges: (i) Aspect Category Extraction (or sometimes Aspect term extraction), and (ii) Polarity Classification. Aspect Term Extraction refers to the task of identifying the categories and terms that relate to those categories within reviews \cite{yin2016unsupervised}. After categories are identified, a polarity needs to be assigned to each one, thus performing the task of polarity classification for a particular aspect. In general terms the techniques used are either machine learning related or lexical techniques \cite{kastrati2020weakly}. Machine learning based techniques perform classification based on supervised algorithms which make use of annotated data, or unsupervised (weakly supervised) algorithms which make use of a few or no annotated data. We have to keep in mind, though, that sometimes hybrid approaches, which make use of both techniques are identified very often in practice. Most of the reviewed papers are either based on machine learning techniques as the only choice \cite{dos2014deep, zhang2019aspect}, or sometimes blend in with lexical techniques as well \cite{ilmania2018aspect}. Since machine learning based techniques have proven to be more successful and are more abundant as part of research experience, this section focuses on those techniques only and doesn't take under consideration the techniques that are based on lexical rules alone. Based on the review of the work in this field, neural networks seem to dominate as a preference of most authors in their work. In \cite{kastrati2020weakly}, the authors use a Deep Network Model, or more specifically a Convolutional Neural Network to learn the aspect categories and label the propagation of aspects into further steps. Convolutional Neural Network are augmented with gating mechanisms in \cite{xue2018aspect}, thus presenting a new approach so called Gated Convolutional Network with Aspect Embedding (GCAE), which the authors claim to be much simpler than recurrent network-based models. Nevertheless, techniques based on Recurrent Networks have been quite abundant. For example, authors in \cite{hazarika2018modeling} have used a Recurrent Network for simultaneous classification of all aspects in a sentence along with temporal dependency processing of their corresponding sentence representations. In \cite{nguyen2015phrasernn}, authors have proposed a new approach of identifying the sentiment of an aspect of an entity through an extension of a Recursive Neural Network (RNN) that takes both dependency and constituent trees of a sentence into account. In addition to Deep Networks, techniques based on Long Short-Term Memory (LSTM) have gained popularity among machine learning solutions. In \cite{ma2018targeted}, authors have augmented a Long Short-Term Memory network with a hierarchical attention mechanism consisting of target-level attention and sentence-level attention. Moreover, in \cite{ruder2016hierarchical} authors have used a hierarchical Bidirectional Long Short-Term Memory Network to model the interdependencies of sentences in a review, thus demonstrating the truthfulness of their initial hypothesis which stated that sentences in a review built upon each other, thus the knowledge of the review structure and sentential context should inform the classification of each sentence. A Bidirectional LSTM has also proven to be useful for a language-agnostic approach proposed in \cite{akhtar2019language}, a network that was further assisted with extra handcrafted features. In addition to network-based approaches, other algorithms like Support Vector Machine (SVM) and Naive Bayes (NB) have proven to be surprisingly effective and accurate. In \cite{ahmad2017sentiment} SVM is used to classify the sentiment in product reviews, while authors in \cite{mubarok2017aspect} have approached the same problem using NB as the algorithm doing the job. Finally, a somewhat different approach has been seen in \cite{sun2019utilizing}, where authors have made use of BERT to construct auxiliary sentences from the aspect and convert Aspect-Based Sentiment Analysis into a sentence-pair classification task. When focusing our attention to education domain research, we can see that in \cite{itani2018understanding} a supervised machine learning based drop-out prediction that uses predictive algorithms (Random Forest and Gradient Boosting) was designed as an automated intervention solution. While, for personalized intervention solutions, Explicative algorithms (Logistic Regression and Decision Tree) are used. Further, in \cite{whitehill2015beyond} the focus was on time invariant classifiers, i.e. classifiers whose input/output relationship is the same the whole time. More specifically, multinomial logistic regression (MLR) with a rigid term on every feature except for the bias term was used for classification.  

The models have gone under the evaluation process once the experiments were conducted. The F-1 score evaluation metric is used very often, although other metrics such as recall, precision, and accuracy have been used as well. 
In the next section, is presented the word embeddings and semantics in ABSA, and with this, we complete our deeper investigation into the techniques used in a number of research papers towards the challenge of Aspect-Based Sentiment Analysis.

\section{Word Embeddings and Semantics in Aspect-Based Sentiment Analysis }
\label{sec:word-embedding}

Word embeddings are a type of word representation that allows words with similar meaning to have a similar representation \cite{karani2018introduction}. In Natural Language Processing (NLP) word embeddings are one of the key breakthroughs and main factors for the impressive performance of deep learning methods on challenging problems, thus the use in ASBA is seeing a significant rise. The reason why word embeddings are a good representation of deep learning methods is their dimensionality. The majority of neural networks do not play well with high-dimensional, sparse vectors \cite{goldberg2017neural}, and word embeddings happen to be low-dimensional and dense. Thus, technically speaking, word embeddings are a class of techniques where individual words are represented as real-valued vectors in a predefined vector space \cite{karani2018introduction}. The use of word embeddings is present in many of the reviewed papers and has played a very important role in the performance of the proposed techniques and frameworks. In \cite{kastrati2020wet}, a dataset containing word embeddings and document topic distribution vectors generated from MOOC video lecture transcripts was presented. The transcripts were taken from 12,032 video lectures from 200 different courses in Coursera. Further, wrod2vec and Latent Dirichlet Allocation (LDA) were used to generate word embeddings and topic vectors, respectively. In \cite{kastrati2019impact}, financial documents classification mechanism was presented. The framework was split into two modules: (i) a documents representation module, and (ii) a classification module. For the classification mechanism to show the desired performance, a document was enriched with semantics using background knowledge provided by an ontology and through the acquisition of its relevant terminology \cite{kastrati2019performance}. This way, in-depth coverage of concepts is achieved and conceptualization is involved in documents captured. Finally, in \cite{kastrati2015semcon, kastrati2016semcon}, an objective metric called SEMCON, was proposed. It is used to enrich existing concepts in domain ontologies for describing and organizing multimedia documents \cite{kastrati2019integrating}. More specifically Part of Speech (POS) is extracted from a morpho-syntactic analysis performed on partitioned messages collected from a document. These and many other pieces of research have shown great importance in incorporating semantical and syntactical techniques in the text representation. In the next section, a general conclusion is drawn and the learned lessons are depicted. 

\section{Conclusion and Lessons Learned}
\label{sec:conclusion}

In this section, a short, reasonable, and conclusion will be made based on the facts extracted from reviewing previous and present work in Aspect-Based Sentiment Analysis. It can be said that Aspect-Based Sentiment Analysis is a relatively new field of study and much future work is needed to reach the desired level. Most of the approaches reviewed have stated that their results, despite being satisfactory, are still not where they should be. One of the problems acting as a bottleneck in further advancement is the lack of large annotated datasets. Another thing to keep in mind is that no technique could be used to ensure good performance in all domains, and each domain required its own set of practices and strategies to reach the desired results. The latest works are being focused on neural network-based solutions, even though some works in which large annotated datasets were missing, other solutions, such as SVM or NB were also successfully utilized. In addition to the algorithms and models chosen, other parts of the process turned out to be just as important, especially preprocessing steps. With this paper, a comprehensive review on Aspect-Based Sentiment Analysis (or even Text/Opinion Mining in general) is given, a review that may serve useful for future approaches to the problem.

\bibliographystyle{unsrt}  
\bibliography{absa}  



\end{document}